\documentclass[sn-mathphys]{sn-jnl}
\usepackage[utf8]{inputenc}
\usepackage{subcaption}

\begin{document}

\title{Distinguishing Risk Preferences using Repeated Gambles}

\author*[1]{\fnm{James} \sur{Price}}\email{james.price.2@warwick.ac.uk}

\author[2,3]{\fnm{Colm} \sur{Connaughton}}\email{c.connaughton@lml.org.uk}

\affil*[1]{\orgdiv{Centre for Complexity Science}, \orgname{University of Warwick}, \orgaddress{\city{Coventry}, \country{United Kingdom}}}

\affil[2]{\orgname{London Mathematical Laboratory}, \orgaddress{\city{London}, \country{United Kingdom}}}

\affil[3]{\orgdiv{Mathematics Institute}, \orgname{University of Warwick}, \orgaddress{\city{Coventry}, \country{United Kingdom}}}

\abstract{Sequences of repeated gambles provide an experimental tool to characterize the risk preferences of humans or artificial decision-making agents. 
The difficulty of this inference depends on factors including the details of the gambles offered and the number of iterations of the game played.
In this paper we explore in detail the practical challenges of inferring risk preferences from the observed choices of artificial agents who are presented with finite sequences of repeated gambles.
We are motivated by the fact that the strategy to maximize long-run wealth for sequences of repeated additive gambles (where gains and losses are independent of current wealth) is different to the strategy for repeated multiplicative gambles (where gains and losses are proportional to current wealth.)
Accurate measurement of risk preferences would be needed to tell whether an agent is employing the optimal strategy or not.
To generalize the types of gambles our agents face we use the Yeo-Johnson transformation, a tool borrowed from feature engineering for time series analysis, to construct a family of gambles that interpolates smoothly between the additive and multiplicative cases. 
We then analyze the optimal strategy for this family, both analytically and numerically. 
We find that it becomes increasingly difficult to distinguish the risk preferences of agents as their wealth increases. 
This is because agents with different risk preferences eventually make the same decisions for sufficiently high wealth. 
We believe that these findings are informative for the effective design of experiments to measure risk preferences in humans.}

\maketitle

\section{Introduction \& Motivation}

People, organisations, algorithms and other autonomous agents must often make decisions in circumstances where there are uncertainties about the possible outcomes. 
Different agents may rank possible actions differently according to the uncertainties involved. 
For example, one person may prefer to keep a stable job guaranteeing a modest income rather than quit to join a start-up which might yield a very high income or none at all depending on whether it succeeds or fails. 
A second person might make the opposite decision. 
We say that these people have different risk preferences: the first is risk averse, the second is risk-seeking. 
Quantifying risk preferences is important for understanding human \cite{peterson2021using}, organisational \cite{cramer2002low} and algorithmic decision making \cite{grote2020ethics}. 
Expected utility theory is a common approach to modelling differences in risk preferences \cite{o2018modeling}. 
The basic idea is that each agent possesses an intrinsic utility function that numerically encodes the abstract value of different outcomes. 
For decisions where the outcomes of different choices are certain, an agent chooses the option yielding the greatest change in its utility. 
For decisions where the outcomes of different choices are uncertain, the agent calculates the probability-weighted sum of possible changes in its utility - the so-called ``expected utility change'' - and chooses the option yielding the greatest expected utility change. 
This model assumes that the intrinsic utility functions of all agents and the probabilities associated with different options are known in advance.

Gambles can used to probe risk preferences experimentally \cite{charness2013experimental,lichtenstein1971reversals,holt2002risk}. 
In such experiments, an agent is required to choose between gambles that involve different levels of risk and reward with probabilities that are known a-priori. 
By recording which gambles the agent selects, an observer should be able infer information about the risk preferences of the agent and thereby estimate the agent's utility function. 
Identifying the risk preferences of expected utility maximizing agents is the aim of Samuelson's Theory of Revealed Preferences \cite{samuelson1938note}.
This theory follows the argument outlined above that repeatedly observing an agent's choices provides information about their risk preferences.
Building on this Afriat showed in \cite{afriat1967construction} that under a fairly general set of conditions it is possible when given a finite sample of preference data to construct a utility function that is consistent with that behaviour.

In what follows, we make a distinction between different types of gamble an agent may be presented with. 
In the first instance we distinguish between additive and multiplicative gambles. 
For additive gambles, gains and losses are independent of an agent’s current wealth. 
For multiplicative gambles, gains and losses are proportional to an agent’s current wealth.
This distinction only becomes relevant in context of repeated gambles.
By ``repeated'' we mean that gains and losses are carried forward from one decision to the next.  
For multiplicative gambles, the carrying forward of winnings means that the magnitude of potential gains and losses will grow if an agent is doing well and shrink if an agent is doing badly.
This is not the case for additive gambles.
The distinction has important consequences for the question of how an agent's risk preferences affect overall performance in long sequences repeated gambles.
It is known from the work of Kelly \cite{kelly1956new} that optimal risk preferences differ for repeated additive versus multiplicative gambles. 
For additive gambles it is long-run optimal to choose gambles that maximise the expected change in wealth. 
In contrast, for multiplicative gambles, it is long-run optimal to choose gambles that maximise the expected change in the logarithm of wealth.
The implication is that for additive gambles, agents with a linear utility function should do best in the long run whereas for multiplicative gambles, agents with a logarithmic utility function should do best in the long run.
This implication can be made precise using the notion of a long-term growth rate for agent's wealth which we explain in Section~\ref{sec1}.
In this paper, we construct and explore a family of gambles that interpolates continuously between the additive and multiplicative cases. 
This is done in Section~\ref{sec2}.
Placing the additive and multiplicative cases on a spectrum helps to put the differences between them into a context. 
It also allows us to determine how the choice of gamble type affects the ability to infer agents' utility functions from observations.
We find that within this family, it generally becomes increasingly difficult to distinguish the risk preferences of agents as their wealth increases. 
This is because agents with different risk preferences eventually make the same decisions for sufficiently high wealth.
This is explained in Section~\ref{sec3} where we provide a detailed description
and analysis of our numerical experiments.
The results of these experiments, and the inference of utility functions in various different contexts are presented in Section~\ref{sec4}.
We conclude with some discussion of the implications of these results for real experiments on measuring risk preferences and possible avenues for further investigation.

\section{Time-Optimal Strategies for Repeated Gambles} \label{sec1}
A gamble is an action taken by an agent that has an uncertain outcome. 
We are interested in scenarios where an agent is repeatedly presented with a choice of gambles, decides their preferred option and observes it's outcome.
The outcome of each gamble is added to the wealth of the agent, therefore the wealth represents the accumulated reward.
We assume the agent has perfect information about the structure of uncertainty for each gamble and therefore the wealth of an agent can be modelled as a random process.

An important question is how an agent decides preferences between gambles, the answer depends on what the agent is trying to maximize.
Our agents are wealth-maximizing however directly maximizing the observed reward from a gamble is impossible because as the outcomes are uncertain we only know the reward after the gamble has been chosen.
Instead we could choose to prefer the gamble with the highest expected reward, that is, if we could replay the outcome of every gamble many many times and take the average, which gamble would have the highest average.
However, in some cases this approach is misleading because low probability, high impact events can make the expected reward much larger than the typical reward.
If the size of the gamble payout depends on an agent's current wealth then by the time an agent gets lucky enough the gamble size is much smaller, this is the principle of the coin toss experiment explained in \cite{peters2019ergodicity}.

An alternative approach that avoids this pitfall is to maximize the long-term growth rate as outlined in \cite{peters2018time}.
The growth rate of a random process, $X_t$, which has independent and identically distributed (iid) increments is defined as,
\begin{equation}
    \lim_{T \rightarrow \infty} \frac{ X_T-X_0}{T}
\end{equation}
Growth-rate maximization allows us to invoke the concept of time-optimality.
If an agent is behaving time-optimally then they are guaranteed at the infinite-time limit to have a larger wealth than any other agent.
Time-optimality is a useful condition to have because the outcomes of individual gambles are uncertain however we can still provide a guarantee that the growth-rate maximizing agent will do better than all other agents.
The reason behind this guarantee is that the observed average increment size converges to the theoretical average increment size in the infinite-time limit.
Therefore the agent with the largest theoretical average increment size, that is the agent with the largest growth rate, will have the largest realised wealth.
This is a restatement of the Strong Law of Large Numbers (SLLN).

We now calculate the growth rate for a couple of example random processes.
An additive process is defined by $X_{t+1} = X_t + Z_t$ for some iid sequence of random variables $Z_t$, therefore the average increment size is independent of the current value of the random process. 
To calculate the growth rate we write,
\begin{align}
    \lim_{T \to \infty} \frac{ X_T-X_0}{T} &= \lim_{T \to \infty} \frac{1}{T} \sum_{t=1}^{T} \big( X_t - X_{t -1} \big) \\ &= \lim_{T \to \infty} \frac{1}{T} \sum_{t=1}^{T} Z_{t} = \mathbb{E} ( Z_1 )
\end{align} 
The final equality is an application of the SLLN. 

Multiplicative processes are random processes where the average incremental size is proportional to the current value.
An example of a multiplicative process is $X_{t+1} = Z_t \cdot X_t$ where $Z_t$ is a sequence of iid random variables \cite{redner1990random}.
Multiplicative processes have no directly computable growth rate because the increments $X_{t+1}-X_t=(Z_t-1)X_t$ are not iid.
Instead we rewrite the original process as,
\begin{equation} \label{eq14}
    \log X_{t+1} = \log Z_t + \log X_t
\end{equation}
which is an additive process.
From here we calculate the growth rate of $\log X_t$ and use this as a proxy for the growth rate of $X_t$.
In this case the growth rate is given by $\mathbb{E} ( \log Z_1 )$. 

The approach of transforming our random process to one with nicer properties is well-known in time series analysis and is basis of methods such as the Box-Cox transform \cite{box1964analysis}.
The purpose of the Box-Cox transform is to transform a time-series, using a so-called power transform, into a white-noise process.
White-noise processes have many desirable properties that make them suitable for a wide range of statistical methods.
Similarly in growth-rate maximization a random process, which can just be thought of as a theoretical time-series, is first transformed and then inferences are made using the desirable properties of the transformed process.
In growth-rate maximization the analogue of the power transform is called the ergodicity transform.
Hence we can say that for multiplicative processes the ergodicity transform is the logarithmic function because for multiplicative processes, as shown in the previous example, we calculate the growth rate of $\log X_t$.

An alternative to growth-rate maximization is expected utility theory \cite{von1944theory}.
Expected utility theory assigns each wealth value a `utility' that represents the amount of happiness/fulfillment/pleasure an agent feels when they gain this wealth amount.
The function that outputs the utility at a given wealth is called the utility function and utility-maximizing agents therefore maximize $\mathbb{E} \big ( \Delta U(X) \big)$ where $X$ is the payoff from a gamble and $ \Delta U$ is the change in the utility function from an agents current wealth to their future wealth.
Expected utility theory however does not supply a reason to why a specific utility function should be chosen - agents maximize a utility function but expected utility theory can neither predict nor explain the choice in utility function.
The expected utility maximization expression above tells us that growth rate maximizing can be thought of as expected utility maximization with a known utility function.
In multiplicative dynamics the utility function is the logarithmic function and in additive dynamics the utility function is the identity function.
In general if an expected utility maximizing agent is using the ergodicity transform as their utility function then they are time-optimal.

Growth-rate maximization has an established history. 
It was first introduced in it's current form in \cite{peters2016evaluating}, however it already had a home in sports betting and finance in which it was known as the Kelly Criterion \cite{kelly1956new}.
The Kelly Criterion is a formula for calculating the optimal bet size as a proportion of wealth when the winning probability of a bet is known.
Kelly's definition of optimal matches our definition of time-optimal in that for repeated bets an optimal agent will have a higher wealth than all other agents with probability one.
In Kelly's setup an agent bets a percentage of their current wealth and receives a doubled amount back with probability $p$ and loses it otherwise.
This is an alternative way of describing multiplicative dynamics as the reward size is proportional to the current wealth of an agent.
Kelly states that for multiplicative processes it is the expected logarithm of the increments that need to be maximized in order to achieve time-optimality, this matches the conclusion of growth-rate maximization.
The Kelly Criterion is therefore a restricted version of growth-rate maximization in that it only considers multiplicative processes.
Additionally, the types of gambles considered by Kelly are restricted to doubling or losing your bet, and therefore further restricts us to scenarios where there are only two possible outcomes.

\section{Yeo-Johnson Gambles: Interpolating between Additive and Multiplicative Gambles} \label{sec2}
So far we have only examined the scenario in which an agent takes gambles in additive or multiplicative processes.
We saw how each dynamic, additive and multiplicative, is represented by an ergodicity transform.
For additive dynamics the ergodicity transform is the identity function and for multiplicative dynamics the ergodicity transform is the logarithmic function.
This suggests that by considering an ergodicity transform that is `in-between' the identity function and the logarithmic function we can reach random processes that are between additive and multiplicative dynamics.

This idea of being `in-between' the identity and logarithmic functions again links to the methodology of the Box-Cox transform.
As mentioned previously, the aim of the Box-Cox transform is to determine the function that best transforms a given time series into a Gaussian white noise process.
The choice of functions in the Box-Cox methodology is restricted to the power transform family of functions. 
This family of functions has several other names including the Isoelastic Utility function and the Constant Relative Risk Aversion (CRRA) function \cite{arrow1971theory} but importantly for our use it contains both the identity and logarithmic functions.
Each member of this family is represented by a unique value of a single continuous parameter $\gamma$, $\gamma=0$ is the identity function (additive dynamics) and $\gamma=1$ is the logarithmic function (multiplicative dynamics).
As $\gamma$ is a continuous parameter, choosing $\gamma \in (0,1)$ represents a function that, when used as an ergodicity transform, will correspond to a random process with dynamics that are between additive and multiplicative.
When $\gamma$ is close to $0$ we have a random process that is close to additive dynamics and when $\gamma$ is close to $1$ we have a random process that is close to multiplicative dynamics.
These functions are defined as,
\begin{equation}
\Psi_\gamma(X_t) = 
\begin{cases}
\frac{X_t^{1-\gamma}-1}{1-\gamma} \quad \text{for } \gamma \neq 1\\
\log{X_t} \quad \text{for } \gamma = 1
\end{cases}
\end{equation}
There is an immediate issue with using this family of functions.
In the parameter range $\gamma \in (0,1)$, that is the set of so-called `in-between' functions, the Isoelastic function is bounded below at $x=0$ by $\frac{-1}{1-\gamma}$.
In the case that a random process has large negative values we cannot apply the Isoelastic function in the parameter range $\gamma \in (0,1)$ as this results in a complex-valued outcome.
From this explanation we can conclude that for a transform function to be suitable for our purposes it must have an unbounded codomain.

\begin{figure}[h]
\centering
\includegraphics[width=0.6\linewidth]{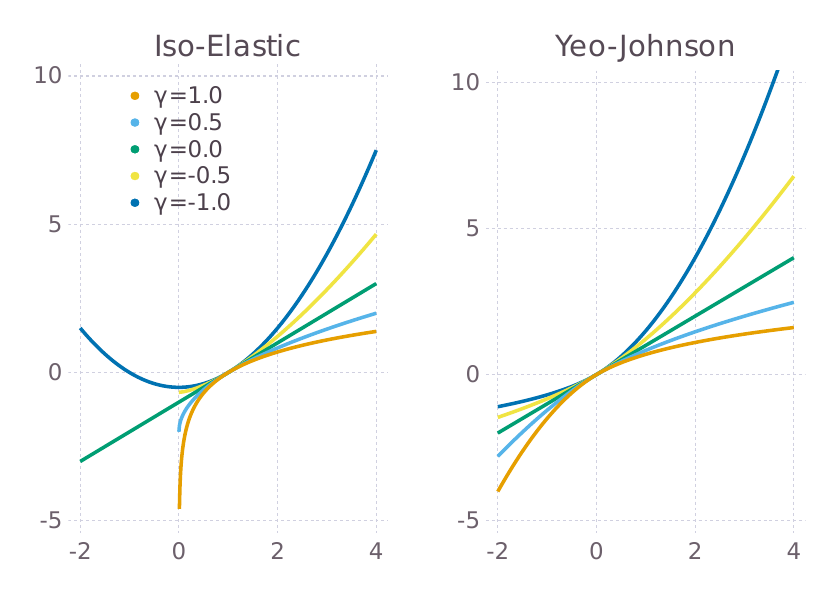}
\caption{An illustration of the difference between the Isoelastic and Yeo-Johnson transforms.}
\label{fig3}
\end{figure}

A family of functions that satisfies this criteria is an adaptation to the Isoelastic transform called the Yeo-Johnson transform \cite{yeo2000new}, defined as,
\begin{equation} \label{eq2}
\Psi_\gamma(X_t) = 
\begin{cases}
\frac{(X_t+1)^{1-\gamma}-1}{1-\gamma} \quad \text{for } \gamma \neq 1, X_t \geq 0\\
\log{(X_t+1)} \quad \text{for } \gamma = 1, X_t \geq 0 \\
\frac{1-(1-X_t)^{1+\gamma}}{1+\gamma} \quad \text{for } \gamma \neq -1, X_t < 0\\
-\log{(1-X_t)} \quad \text{for } \gamma = -1, X_t < 0 \\
\end{cases}
\end{equation}
Note that we have shifted the parameter $\gamma$ from how it was stated in the original paper using the substitution $\lambda = 1-\gamma$, this was done to better align with the parameter from the Isoelastic transform.
The Yeo-Johnson transform is illustrated alongside the Isoelastic transform in Figure~\ref{fig3}. 
The benefit of using the Yeo-Johnson transform is that it is unbounded for $\gamma \in [-1,1]$, a much wider range of parameter values than for the Isoelastic transform. 
As the Yeo-Johnson transform is strictly increasing this means that for $\gamma \in [-1,1]$ the inverse Yeo-Johnson transform, $\Psi^{-1}_\gamma (x)$, is well-defined.
Further the second derivative of the Yeo-Johnson transform at $x=0$ is continuous which results in a smooth transition between negative and positive values.
The identity function, and therefore additive dynamics, are part of the Yeo-Johnson family of functions however $\gamma=1$ no longer represents a logarithmic transform. 
Instead we have a transform which, for positive wealth, is multiplicative in $X_t+1$ and therefore asymptotically multiplicative as $X_t\rightarrow \infty$.
This is a much less significant drawback than the singularities from using the Isoelastic function and therefore to categorise the dynamics between additive and multiplicative we will use the Yeo-Johnson transform.

\section{Design Considerations for Revealed Preference Experiments using Repeated Yeo-Johnson Gambles} \label{sec3}
The aim of this work is to design an experiment such that we can identify the growth-rate maximizing agent from a group of in-silico expected utility maximizing agents with each agent having a unique utility function.
As mentioned previously, the growth-rate maximizing agent can be thought of as an expected utility maximizing agent whose utility function is the ergodicity transform.
The growth-rate maximizing agent is time-optimal, meaning that they are guaranteed at the infinite-time limit to have a larger wealth than any other agent.
As a result the longer the experiment runs for, that is the more times an agent has to choose between gambles, the more likely the growth-rate maximizing agent is to have the highest wealth. 
Therefore this experiment attempts to answer whether, in a reasonable time frame, the growth-rate maximizing agent can be identified.
This experimental design work has implications for real-world experiments such as \cite{meder2021ergodicity}, where the risk preferences of participants need to be quantified.

The specific details of the experiment are as follows.
We consider a selection of expected utility maximising agents, $A_i$, whose utility function is given by $U(x)=\Psi_{\eta_i}(x)$, the Yeo-Johnson transform as defined in \eqref{eq2}, for some $\eta_i \in [0,1]$. 
At every time-step all agents have to decide between two gambles, the outcome of which will effect their wealth, $X_{i,t}$. 
All agents see the same set of gambles at every time-step, these gambles are:
\begin{align}
    \textbf{Safe Option:}  \quad X_{i,t+1} &= \Psi^{-1}_\gamma \big( \Psi_\gamma(X_{i,t}) + \lambda_t \big) \label{safe_option}\\
    \textbf{Risky Option:}  \quad X_{i,t+1} &= \Psi^{-1}_\gamma \big( \Psi_\gamma(X_{i,t}) + \pi_{t} \big) \label{risky_option}
\end{align}
The value of $\lambda_t$ in \eqref{safe_option} is uniformly distributed but pre-determined, hence the agent is certain of the outcome of this option. 
We can think of the safe option as placing wealth into a bank account where the future changes in wealth are known and certain.

The value $\pi_t$ in \eqref{risky_option} is a realisation from $\Pi$, a normally distributed random variable with mean $\mu$ and variance $\sigma^2$. 
The parameters $\mu$ and $\sigma$ are fixed and known to the agents for the entire experiment. 
We can therefore think of the risky option as placing wealth into a speculative market where we have perfect information on the uncertainty of future change in wealth. 

The parameter $\gamma$ is fixed at the beginning of the experiment and represents the dynamics of the rewards from the gambles.
However as we are only interested in random processes between additive ($\gamma=0$) and multiplicative ($\gamma=1$) we restrict $\gamma$ to the range $[0,1]$.

To explain the evolution of wealth as described in \eqref{safe_option} and \eqref{risky_option} we can refer to the example of multiplicative processes in Section~\ref{sec1}.
We can rewrite the evolution of wealth under multiplicative dynamics, given by \eqref{eq14}, as,
\begin{equation} \label{eq15}
    X_{t+1} = \exp \big( \log(X_t) + \log(Z_t) \big)
\end{equation}
By noting that $\exp(x) = \log^{-1}(x)$ we can see that \eqref{safe_option} and \eqref{risky_option} are generalised versions of \eqref{eq15} with different payoffs and where the ergodicity transform is from the Yeo-Johnson family.

\subsection{Utility Calculation}
The agents in our experiment make decisions by calculating the expected change in utility of both gambles and choosing whichever option has the highest change.
Every agent has a unique utility function that comes from the Yeo-Johnson family of functions, therefore we can use a single parameter $\eta_i$ to characterize their risk preferences. 
The lower the value of $\eta_i$ the more risk-seeking an agent is.

When an agent chooses the safe option there is complete certainty in the payoff of the gamble.
This means that the expected change in utility is the same as the change in utility.
The utility of an agent with wealth $X_t$ at time $t$ and utility function $U(x)=\Psi_\eta(x)$ with wealth dynamics $\gamma$ is given by,
\begin{align}
    X_{t+1} &= \Psi^{-1}_\gamma \big( \Psi_\gamma(X_t) + \lambda_t \big) \label{eq10} \\
    \implies \Psi_\eta(X_{t+1}) &= \Psi_\eta \Big( \Psi^{-1}_\gamma \big( \Psi_\gamma(X_{t}) + \lambda_t \big) \Big) \label{eq17}
\end{align}
Where $\lambda_t$ is the (known) payoff of the safe option at time $t$.
In this situation $X_t$ is not a random variable but rather a known quantity as it represents the wealth of our agent at time $t$, which is known to the agent when they have to decide what option to choose next.
Therefore the change in utility of our agent is,
\begin{equation}
    \Psi_\eta(X_{t+1}) -\Psi_\eta(X_t) = \Psi_\eta \Big( \Psi^{-1}_\gamma \big( \Psi_\gamma(X_{t}) + \lambda_t \big) \Big) - \Psi_\eta(X_t)
\end{equation}
For the risky option, due to the non-linear nature of how our agent's wealth evolves there is no tractable expression for the expected change in utility.
Instead we find the probability density function of the utility of wealth at time $t+1$ and use numerical integration to find it's expectation. 
Details are provided in Appendix~\ref{appendixA}.

As mentioned previously, the growth-rate maximizing agent is the agent whose utility function matches the ergodicity transform of the rewards from the gambles.
Using \eqref{eq17} we know the growth-rate maximizing agent prefers the safe option if $\lambda_t>\mu$ and the risky option if $\mu>\lambda_t$, where $\mu$ is the mean of the transformed risky payoff distribution $\Pi$. 
s with growth rate $\lambda_t$ for the safe option and $\mathbb{E}(\Pi)=\mu$ for the risky option.

\subsection{Decision Convergence}
Decision convergence is the property that, for certain combinations of reward dynamics and utility functions, as the wealth of an agent diverges from zero the preferences of that agent converge to the preferences of the growth-rate maximizing agent.

We will illustrate this phenomenon using a simple example.
Let $X_t$ represent the wealth of an agent at time $t$, this agent has to choose between the following two gambles,
\begin{align}
    \textbf{Gamble 1: }X_{t+1} &=
    \begin{cases}
        X_t+10 &\text{with probability }\frac{1}{2} \\
        X_t+100 &\text{with probability }\frac{1}{2}
    \end{cases} \\
    \textbf{Gamble 2: }X_{t+1} &= X_t+x
\end{align}
Suppose that our agent is growth-rate maximizing.
The random process $X_t$ is additive and therefore the utility function of our agent is $U(x)=x$.
The expected change in utility when a growth-rate maximizing agent chooses Gamble 1 is,
\begin{equation}
    \mathbb{E}(X_{t+1}-X_t \vert X_t) = \frac{1}{2} (X_t+10) + \frac{1}{2} (X_t+100) -X_t = 55
\end{equation}
The (expected) change in utility when a growth-rate maximizing agent chooses Gamble 2 is,
\begin{equation}
    \mathbb{E}(X_{t+1}-X_t \vert X_t)=x
\end{equation}
Therefore a growth-rate maximizing agent prefers Gamble 1 over Gamble 2 if and only if $55>x$.

We now consider another agent who has a logarithmic utility function.
The expected change in utility when this agent chooses Gamble 1 is,
\begin{align}
    \mathbb{E}(\log X_{t+1}- \log X_t \vert X_t) &= \frac{1}{2} \log (X_t+10) + \frac{1}{2} \log (X_t+100) -\log(X_t)\\
    &= \log \big( \sqrt{(X_t+10)(X_t+100)} \big) -\log X_t \\
    &= \log \big( \sqrt{(X_t+55)^2-2025}\big) -\log X_t
\end{align}
The (expected) change in utility when this agent chooses Gamble 2 is,
\begin{equation}
    \mathbb{E}(\log X_{t+1}-\log X_t \vert X_t)=\log (X_t + x)-\log X_t
\end{equation}
Therefore a logarithmic utility maximizing agent prefers Gamble 1 over Gamble 2 if and only if $\sqrt{(X_t+55)^2-2025}>X_t+x$.

Decision convergence occurs as the wealth of non growth-rate maximizing agents diverges from zero.
For a logarithmic utility maximizing agent as $X_t \to \pm \infty$ we have $\frac{\sqrt{(X_t+55)^2-2025}}{X_t+55} \to 1$ which means that for large wealth the expressions $\sqrt{(X_t+55)^2-2025}$ and $X_t+55$ are near identical.
Therefore our inequality for when this agent prefers Gamble 1 over Gamble 2 tends to $X_t+55>X_t+x \iff 55>x$. 
This is exactly the same condition as for the growth-rate maximizing agent.
Hence we have shown that as the wealth of the logarithmic utility maximizing agent diverges from zero their preferences align with the preferences of the growth-rate maximizing agent.

Whether or not decision convergence occurs depends on two factors: the utility function of the agent and the ergodicity transform of the rewards from gambles.
In Appendix~\ref{appendixB} we show that this effect occurs for most parameter values for Yeo-Johnson gambles.
Specifically when the utility function of the agent and ergodicity function of the rewards from gambles are the Yeo-Johnson transform with parameters $\eta$ and $\gamma$ respectively then decision convergence occurs for $\eta \in [0,1]$ and $\gamma \in [0,1)$.
Note that if $\gamma= 1$ (multiplicative dynamics) then decision convergence does not occur for any $\eta \in [0,1)$.

The aim of our experiment is to distinguish the risk preferences of agents.
Decision convergence makes distinguishing risk preferences difficult because as the experiment progresses the wealth of agents will diverge from zero and their risk preferences will all converge.
It is impossible to distinguish risk preferences if all agents, regardless of utility function, make the same decisions.
The solution to this issue is to choose parameters for the gamble rewards such that the wealth of agents does not stray too far from zero. 
This further motivates our use of the Yeo-Johnson transform as a suitable function for interpolating between additive and multiplicative dynamics as it is essential for our experimental setup to accommodate negative wealth.

\subsection{Choosing Good Growth Rates}

In order to avoid the effects of decision convergence we will choose the distribution of rewards from gambles such that the growth rate of the time optimal agent is zero.
To find the growth rate of the time optimal agent we first make the modelling decision that $\lambda_t \sim \text{Unif}(\mu-c \sigma^2,\mu+c \sigma^2)$ for some $c \in [0,1]$ and where $\mu$, $\sigma$ are mean and variance respectively of $\Pi$, the normally distributed risky option payoff.

In Section~\ref{sec1} we only defined the growth rate for a process with iid increments which is now no longer applicable as the payoffs for the safe and risky options each follow different distributions.
Instead we apply the Kolmogorov Criterion for the SLLN \cite[page 259]{feller1968introduction} which states that for $Z_t$, a sequence of independent random variables, if $\sum_{t=1}^\infty \frac{\mathbb{V}ar(Z_t)}{t^2}$ converges then we can still apply the SLLN.
In our experimental setup $Z_t = U_\gamma(X_{i,t+1}) - U_\gamma(X_{i,t})$ which when substituted into the equation for the risky option \eqref{risky_option} gives us $\mathbb{V}ar(Z_t) = \mathbb{V}ar(\Pi) = \sigma^2$ and when substituted into the equation for the safe option \eqref{safe_option} gives us $\mathbb{V}ar(Z_t) = \mathbb{V}ar(\lambda_t)=\frac{c \sigma^2}{6}$.
As the incremental variance is independent of the time step the summation converges and thus the Kolmogorov Criterion is satisfied.

Now we know the growth rate exists we can calculate its value in terms of the experimental parameters $\mu$, $\sigma$ and $c$.
As $U_\gamma(X_{i,t})$ is an additive process the growth rate represents the average increment size.
The growth-rate maximizing agent prefers the safe option at time $t$ if $\lambda_t>\mu$, hence the payoff distribution when the growth rate maximizing agent prefers the safe option is $\lambda_t \vert \lambda_t>\mu \sim \text{Unif}(\mu, \mu + c \sigma^2)$.
Therefore the growth rate whenever the safe option is chosen is $\mathbb{E}( \lambda_t  \vert \lambda_t> \mu) = \frac{2 \mu + c \sigma^2}{2}$. 
The growth rate of the risky option is always $\mu$ as the reward is only known once the risky option has been chosen.
Finally we have $\mathbb{P}(\lambda_t>\mu) = \mathbb{P}(\lambda_t<\mu)=\frac{1}{2}$.
Using the law of total expectation, the average incremental size and therefore the growth rate of the growth-rate maximizing agent is given by,
\begin{align}
    &\mathbb{P}(\lambda_t>\mu)\mathbb{E}( \lambda_t  \vert \lambda_t> \mu)+ \mathbb{P}(\lambda_t<\mu) \mu \\
    = &\frac{1}{2}\frac{2\mu + c\sigma^2}{2} + \frac{1}{2}\mu
\end{align}
Rearranging gives us the value of $\mu$ that corresponds to the growth-rate maximizing agent having a growth rate of zero as,
\begin{equation} \label{eq7}
\mu = \frac{-c \sigma^2}{4}
\end{equation}

\subsection{Choosing Informative Gamble Pairs} \label{sec21}

Disagreement amongst agents about the preferred gamble is what allows us to distinguish their risk preferences.
Hence a second and related issue to decision convergence is how to choose the bounds of the uniform distribution of $\lambda_t$ such that we have a high probability of disagreement in what our selection of agents thinks is the better gamble.

\begin{figure}[h]
\centering
\includegraphics[width=0.6\linewidth]{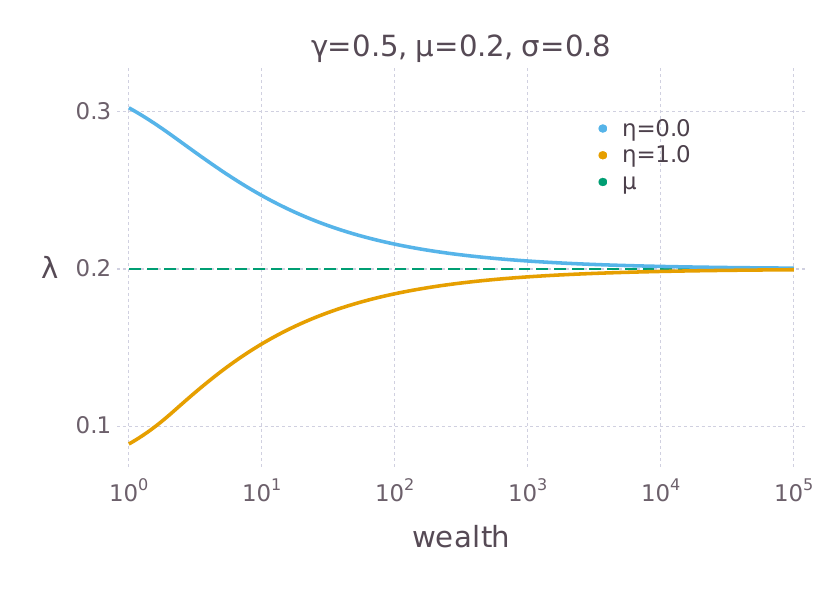}
\caption{A visual illustration of decision convergence: The blue and orange lines represent the value of $\lambda_t$, for agents $A_1$, $A_2$ with $\eta_1=0$ and $\eta_2=1$ such that the expected change in utility of both options are equal.}
\label{fig2}
\end{figure}

Figure~\ref{fig2} shows the risk preference of an agent with $\eta_i=0$ against an agent with $\eta_i=1$. 
The curves represent the value of $\lambda_t$ at a given wealth such that the agent is indifferent between the safe and risky options denoted $\lambda_T$. 
In other words $\lambda_T$, which depends on $X_{i,t}$, satisfies the following equation,
\begin{equation} \label{eq171}
    \mathbb{E} \bigg( \Psi_\eta \Big( \Psi^{-1}_\gamma \big( \Psi_\gamma(X_{i,t}) + \Pi \big) \Big) \bigg) = \Psi_\eta \Big( \Psi^{-1}_\gamma \big( \Psi_\gamma(X_{i,t}) + \lambda_T \big) \Big)
\end{equation}
The left-hand side of \eqref{eq171} represents the expected utility of choosing the risky option and the right-hand side represents the utility from choosing the safe option with known payoff $\lambda_T$.
We can arrange \eqref{eq171} to get an expression for $\lambda_T$ by applying the function $(\Psi_\gamma \circ \Psi_\eta^{-1})(x)$ to both sides and then subtracting $\Psi_\gamma(X_{i,t})$.
This is the expression plotted in Figure~\ref{fig2} for both $\eta=0.0$ and $\eta=1.0$.

If the realisation of $\lambda_t$ is above the line then the agent will prefer the safe option and otherwise they will prefer the risky option. 
The property that the curves converge to $\mu$ is the decision convergence property. 
Disagreement occurs when the realisation of $\lambda_t$ is between the risk preference curves as this means one agent prefers the risky option whilst the other prefers the safe option. 

To create the most informative options we therefore choose the parameter $c$ such that $\lambda_t$ splits the opinions of agents as often as possible in an even manner. 
If we choose the bounds to be too small, the worst and near-worst agents will never disagree with each other. 
Conversely, if we choose the bounds to be too large then the optimal and near-optimal agents will very rarely disagree with each other and due to decision convergence the rate of disagreement will reduce as the wealth of agents increases.

We now present an algorithm to find values of $\mu$ and $c$ that attempt to satisfy the criteria mentioned just above. 
Consider agents $A_p$, $A_q$, who are the agents in our group with the most extreme risk preferences. 
That is, for all agents $A_i$ with risk preferences $\eta_i$ we have $\eta_p \leq \eta_i \leq \eta_q$ where $\eta_p$, $\eta_q$ are the risk preferences of $A_p$, $A_q$ respectively. 
We then find the expected utility of the risky option for $A_p$ and $A_q$ when each agent has zero wealth. 
Due to decision convergence this is precisely the value of wealth where agents are most likely to disagree with each other. 
For agents $A_p$, $A_q$ we then find the value of $\lambda_t$ which results in the utilities of both the risky and safe option being equal at zero wealth and denote these values $\lambda_p$, $\lambda_q$ respectively.
The rationale is that if $\lambda_t<\lambda_p$ or $\lambda_t>\lambda_q$ then all agents, at any wealth, are guaranteed to agree on which option is best. 
We want to minimise the chance of this happening, and hence choose $c$ to be the smallest value such that $[\lambda_p,\lambda_q] \subset [\mu-c \sigma^2,\mu + c \sigma^2]$.

However this approach generates a cyclical problem where $c$ is determined by $\mu$ which in turn is determined by $c$ which is determined by $\mu$ and so on. 
Fortunately numerical studies support the hypothesis that once $\gamma$, $\eta_p$, $\eta_q$, $\sigma$ have been chosen the above recursion converges to a fixed point for the parameter pair $(\mu,c)$.

\begin{figure}[h]
\centering
\includegraphics[width=0.6\linewidth]{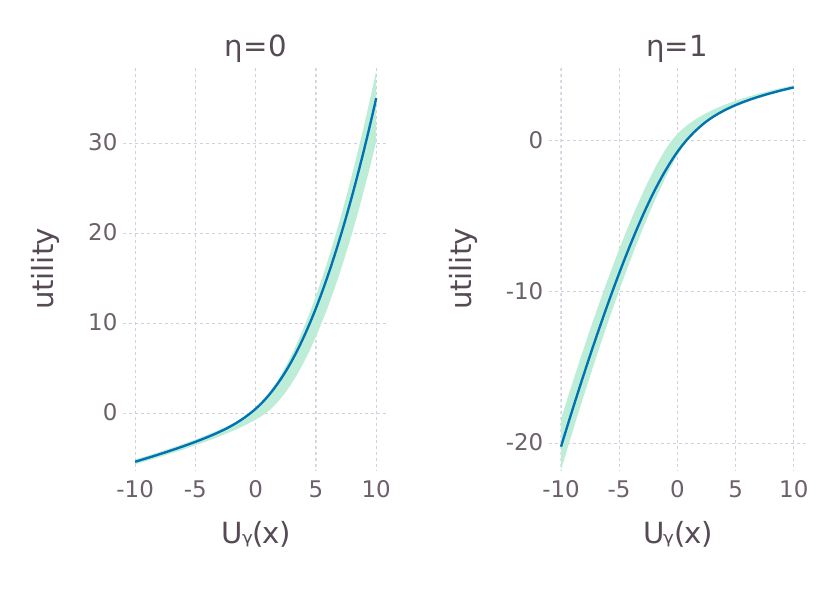}
\caption{A visual illustration of the ideal value of $c$. 
In each subplot the blue line represents the expected utility of the risky option for a given transformed wealth $U_\gamma(x)$ and the green shaded area represents the range of utilities for the safe option corresponding to the range of values for $\lambda_t$. 
The risk preferences of the agents are $\eta=0$ and $\eta=1$.
The values of other parameters are: $\gamma=0.5$, $\mu=-0.166$, $\sigma=2$, $c=0.166$.} 
\label{fig4}
\end{figure}

Figure~\ref{fig4} provides a visual guide for our motivation for the choice of $c$. 
Each subplot represents the risk preferences of agents $A_p$, $A_q$ with $\eta_p=0$ and $\eta_q=1$ respectively. 
The blue line represents the expected utility of the risky option as a function of $\Psi_\gamma(X_t)$ and the green shaded area represents the range of utility for the safe option, which is based on the possible range of $\lambda_t$. 
In both subplots $\mu$ and $c$ are the ideal values outlined earlier in this section. 
At a given wealth if there is any green area above the blue line in the left subplot or below the blue line in the right subplot then this means there is the possibility can all agents agree at a time step. 
We want to minimize the possibility of this happening, which is achieved by keeping $c$ small. 
If there is a gap between the blue line and the green area this suggests no matter the value of $\lambda_t$ an agent will always choose the same option, we also want to minimise the frequency that this occurs.
With our ideal value of $c$, nowhere in our wealth range does the blue line leave the green region. 
This is due to the condition that $[\lambda_p,\lambda_q] \subset [\mu-c \sigma^2,\mu + c \sigma^2]$. 
In fact if $c$ was any smaller we would see one of blue lines leave the green area, this is because $c$ has to be the smallest value such that the above subset condition holds.

\section{Results of Numerical Experiments} \label{sec4}
In this section we show the results of the experiment outlined in Section~\ref{sec3}.
The aim of the experiment was to show that, in a realistic time-period, the behaviour of agents with various risk preferences can be distinguished from one another.

For each experiment we considered a different dynamic.
The dynamics tested were $\gamma \in \{0.0,0.25,0.5,0.75,1.0\}$ as in described in \eqref{safe_option} and \eqref{risky_option}.
In each dynamic we tracked the behaviour of five agents, each with different risk preferences.
The utility functions of our agents came form the Yeo-Johnson family and were controlled by the parameter $\eta$.
The specific parameter values were $\eta \in \{0,0.25,0.5,0.75,1\}$ and was deliberately chosen to match the values of $\gamma$ so that in all dynamics there is a growth-rate maximizing agent.

Other than for the case $\gamma=1$, the parameter values $\mu$ and $c$, representing the mean transformed payoff of the risky option and the range of values of the safe option respectively, were chosen using the algorithm described in Section~\ref{sec21} with the initial choice of $\sigma =2.0$.
Recall that when $\gamma=1$ the decision convergence does not exist for our selection of utility functions and therefore in the $\gamma=1$ dynamic there is no need to ensure a suitable number of informative gambles. 
Instead we use the same parameter choices as for $\gamma=0.75$.
In all dynamics, every agent has an initial wealth of zero.
Every numerical experiment was performed 100,000 times and every experiment had a different sequence of $\lambda_t$.

The sub-figures of Figure~\ref{fig5} show the cumulative distribution function (CDF) of wealth for our five agents under different dynamics and at different points in time.
The left column of Figure~\ref{fig5} shows a snapshot of agent's wealth at $t=30$ and the right column shows a snapshot of wealth at $t=300$.
The specific parameters used for each sub-plot can be found in it's title.

As mentioned above the choice of parameters for $\gamma$ and $\eta$ are equal which means that in every dynamic we can observe the outcome of the growth rate maximizing agent.
This is important for achieving our aim of distinguishing risk preferences.
In each sub-figure of Figure~\ref{fig5} the CDF of the growth-rate maximizing agent is represented by a dashed line.

The concept of time-optimality does not extend neatly into finite-time, we know that the time-optimal agent is guaranteed to eventually surpass all other agents however this surpassing could be on a time-scale that is much longer than our finite-time game allows.
We still expect the growth-rate maximizing agent to perform the best compared to all other agents but  we are hesitant in defining exactly what characterizes the `best' wealth distribution.

One interpretation of the growth-rate maximizing agent is that they are equipped with the utility function that is best suited to the payoff distribution.
However in some cases agents with a lower $\eta$ value have a wealth distribution with a heavier upper tail than the growth-rate maximizing agent.
This is a result of some risk seeking agents getting lucky with their random outcomes and we can see that in the lower tail for the risk seeking agents is significantly smaller than that of the growth-rate maximizing agent.
Therefore we can conclude that for every risk seeking agent who does well there is another equally risk seeking agent who does very poorly.

When comparing between the right and left columns of Figure~\ref{fig5} we can see that there are fewer overlaps between the CDFs of different agents.
This supports our theoretical prediction that the longer the experiment runs for the more prominent the the CDF of the growth-rate maximizing agent would be.

\begin{figure}[!b]]
\centering
\begin{subfigure}{0.48\linewidth}
\centering
\includegraphics[width=\linewidth]{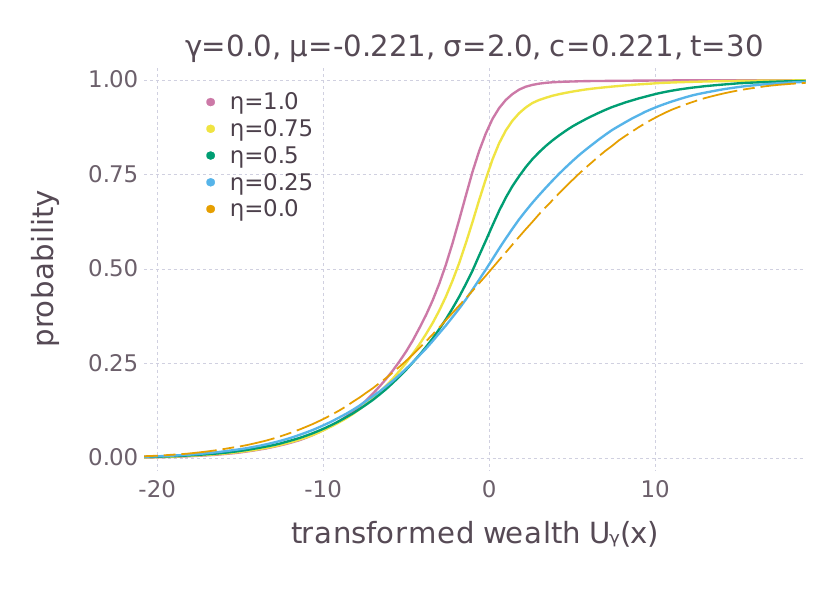}
\end{subfigure}
\hfill
\begin{subfigure}{0.48\linewidth}
\centering
\includegraphics[width=\linewidth]{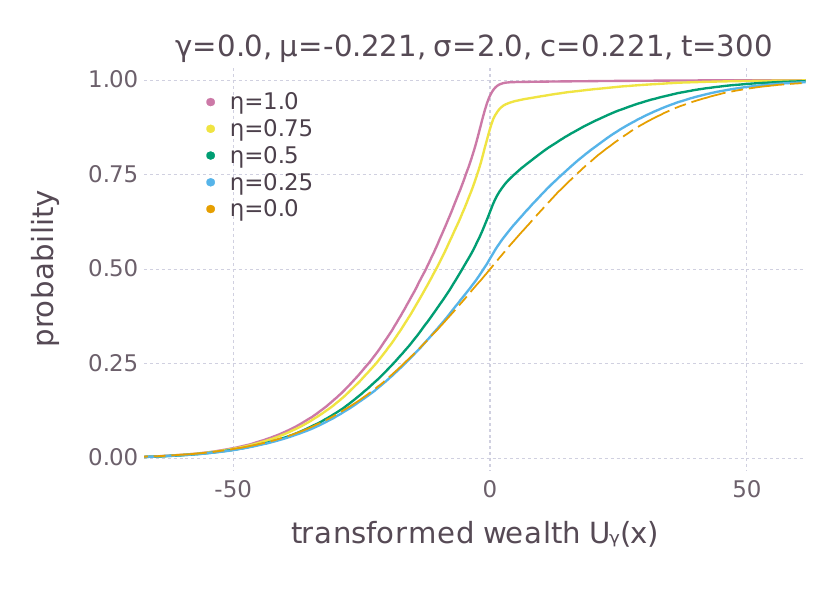}
\end{subfigure}
\hfill
\begin{subfigure}{0.48\linewidth}
\centering
\includegraphics[width=\linewidth]{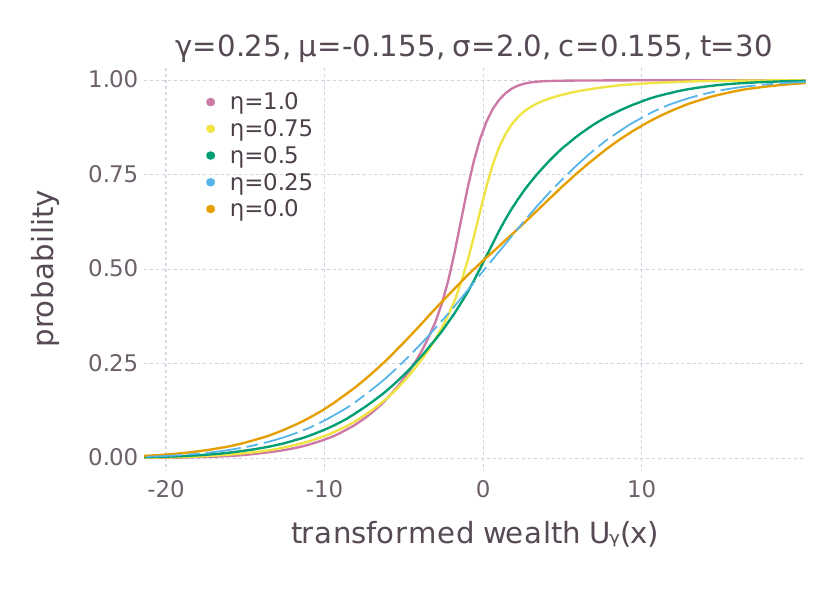}
\end{subfigure}
\hfill
\begin{subfigure}{0.48\linewidth}
\centering
\includegraphics[width=\linewidth]{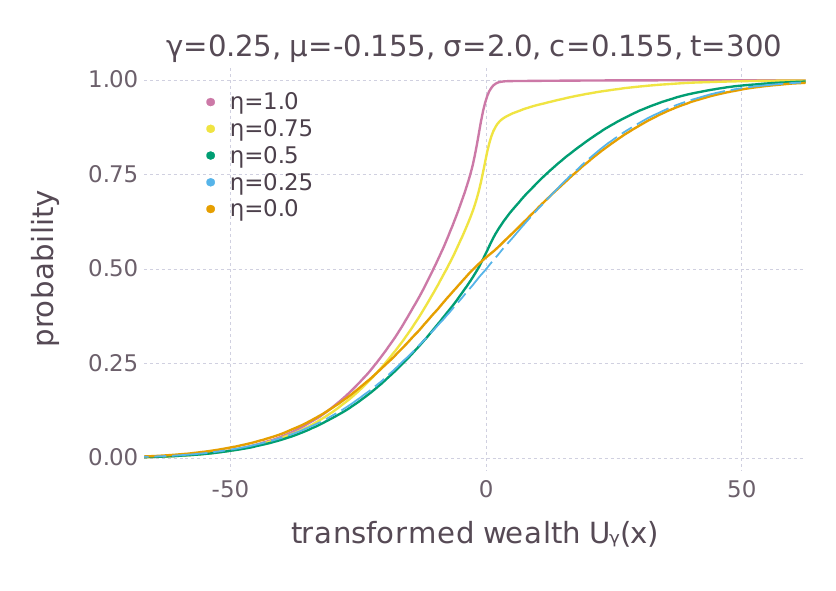}
\end{subfigure}
\hfill
\begin{subfigure}{0.48\linewidth}
\centering
\includegraphics[width=\linewidth]{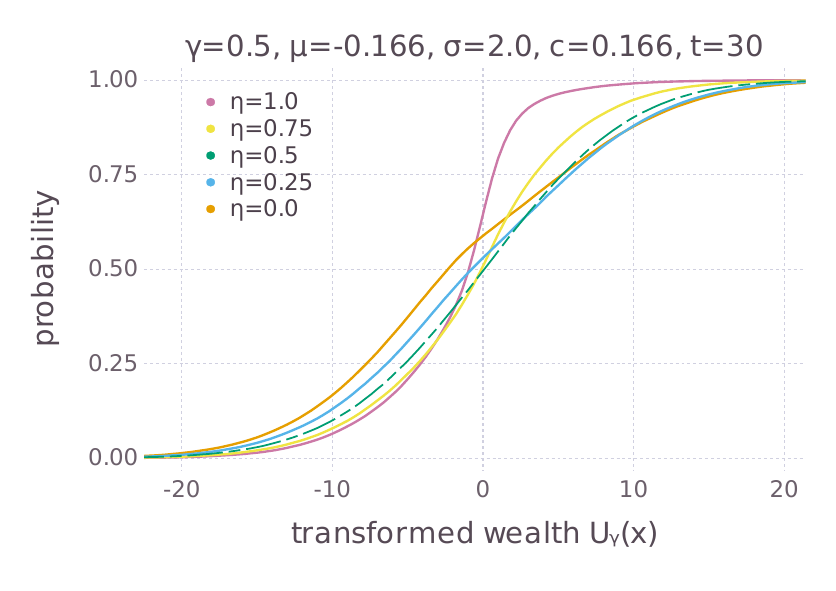}
\end{subfigure}
\hfill
\begin{subfigure}{0.48\linewidth}
\centering
\includegraphics[width=\linewidth]{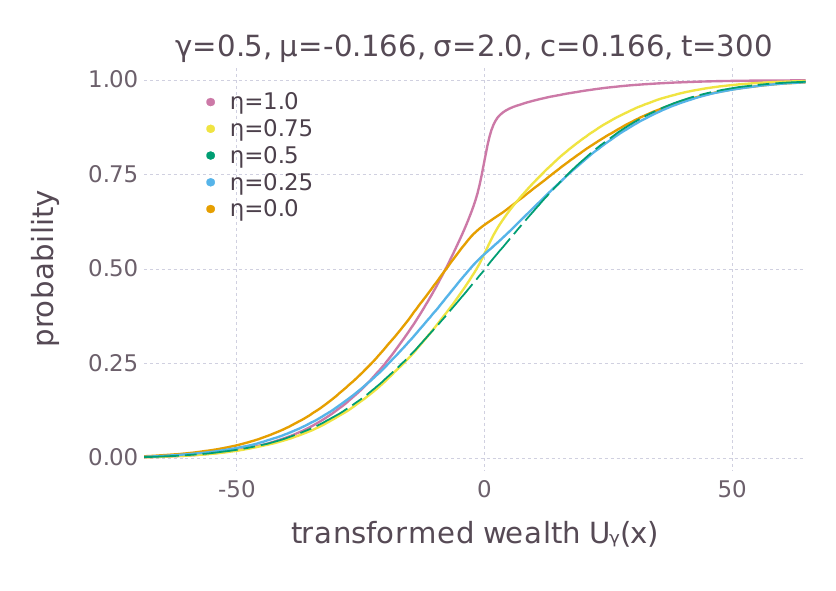}
\end{subfigure}
\hfill
\begin{subfigure}{0.48\linewidth}
\centering
\includegraphics[width=\linewidth]{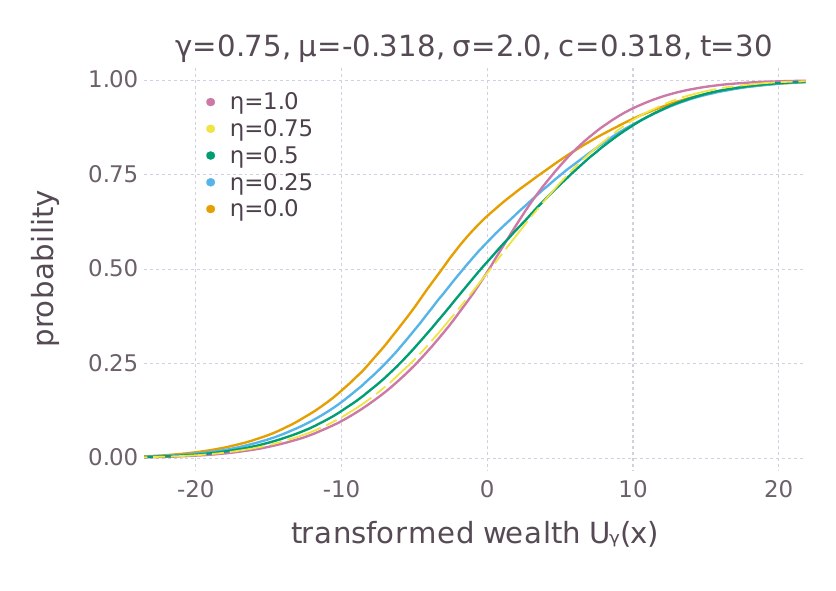}
\end{subfigure}
\hfill
\begin{subfigure}{0.48\linewidth}
\centering
\includegraphics[width=\linewidth]{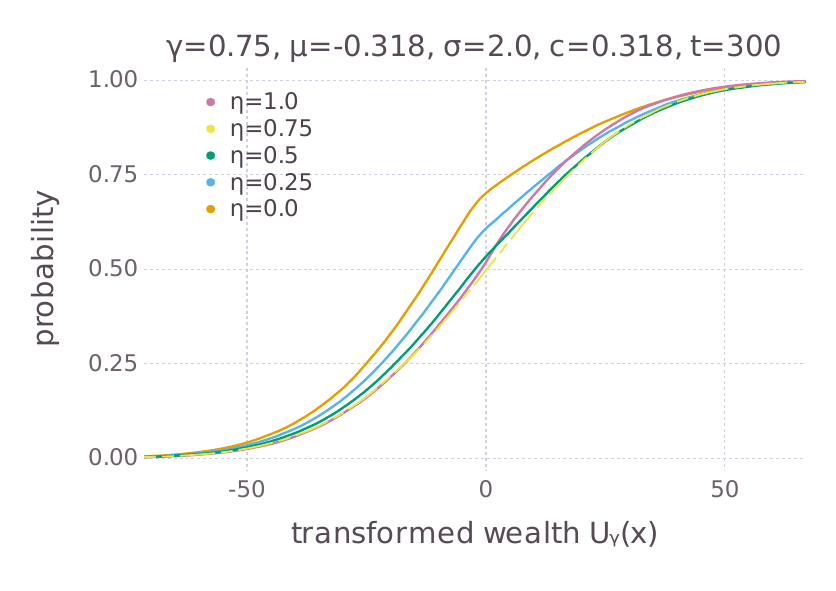}
\end{subfigure}
\end{figure}

\begin{figure} \ContinuedFloat
\begin{subfigure}{0.48\linewidth}
\centering
\includegraphics[width=\linewidth]{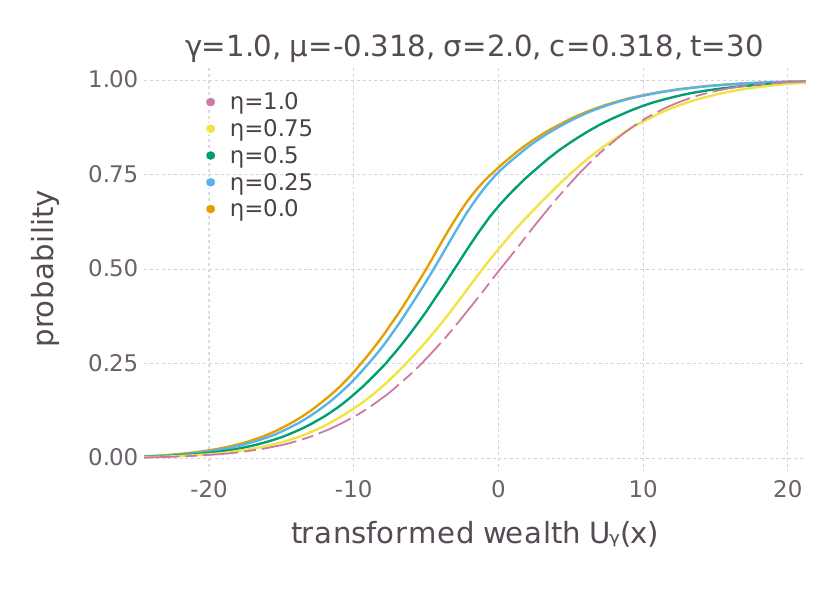}
\end{subfigure}
\hfill
\begin{subfigure}{0.48\linewidth}
\centering
\includegraphics[width=\linewidth]{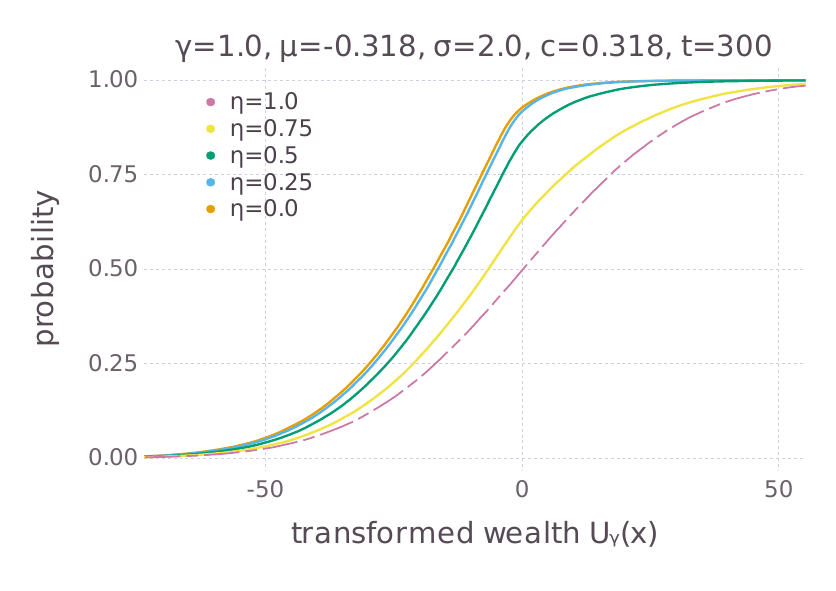}
\end{subfigure}
\caption{Empirical CDFs of agents with different risk preferences under various dynamics. Ideally we would like to see that the set of agents with $\eta_i=\gamma$ have the `best' CDF.}
\label{fig5}
\end{figure}

\section{Discussion and Conclusion}
In this paper we presented a methodology for identifying the risk preferences of in-silico agents.
These agents were placed in a numerical experiment in which they had to repeatedly choose between a safe option and a risky option.
The outcome of these affected their cumulative wealth which was tracked throughout the experiment.
One issue that we encountered was choosing how exactly the outcomes of an agent's decision affected their wealth, the Yeo-Johnson transform was preferred as it is unbounded for a wide range of parameter choices.
Another issue was the so-called decision convergence which describes the phenomenon that as the wealth of agents diverges from zero their risk preferences align. 
This makes distinguishing risk preferences difficult and hence we kept the payoff of the options small so that the wealth of agents stayed around zero.
The outcomes of our numerical experiments were as expected in that the growth-rate maximizing agents performed best.

\appendix
\section{Utility Calculation} \label{appendixA}
In this section we describe how our agents calculate the expected change in utility for the risky option.
We consider agent $A$ with utility function $\Psi_\eta(x)$ in an experiment with dynamics $\gamma$, where $\Psi_\eta(x)$ represents the Yeo-Johnson transform \eqref{eq2} with parameter $\eta$.
First we will calculate the expected change in utility when an agent with wealth $X_t$ at time $t$ chooses the risky option.
The distribution of the utility of wealth when an agent chooses the risky option is given by,
\begin{align}
    X_{i,t+1} &= \Psi^{-1}_\gamma \big( \Psi_\gamma(X_{i,t}) + \Pi \big) \\
    \implies \Psi_\eta(X_{i,t+1}) &+ \Psi_\eta \Big( \Psi^{-1}_\gamma \big( \Psi_\gamma(X_{i,t}) + \Pi \big) \Big)
\end{align}
Where the random variable $\Pi$ which is normally distributed with parameters $(\mu,\sigma^2)$.
The expected change in utility of the agent when the current wealth, $X_t$, is already known is given by the expression,
\begin{equation}
    \mathbb{E} \big( \Psi_\eta(X_{t+1}) - \Psi_\eta(X_t) \big\vert X_t \big) = \mathbb{E} \big( \Psi_\eta(X_{i,t+1}) \big\vert X_t \big) - \Psi_\eta(X_{i,t})
\end{equation}
The value of $\Psi_\eta(X_{i,t})$ is easy to calculate and therefore all that remains to calculate $\mathbb{E} \big( \Psi_\eta(X_{i,t+1}) \big\vert X_t \big)$. 
First we rewrite it as,
\begin{equation}
    \mathbb{E} \bigg( \Psi_\eta \Big( \Psi^{-1}_\gamma \big( \Psi_\gamma(X_{i,t}) + \Pi \big) \Big) \Big\vert X_{i,t} \bigg)
\end{equation}
This expression can be thought of as the expected value of a transformed normal distribution.
The normal distribution is $\Pi$ and the transform function is $g(w)=\Psi_\eta \Big( \Psi^{-1}_\gamma \big( \Psi_\gamma(X_{i,t}) + w \big) \Big)$.
If we denote the distribution of utility at time $t+1$ as $Y$ then,
\begin{equation}
    f_Y(y) = \Big\vert \frac{dg}{dw} \Big\vert f_{\Pi}(w) = \Big\vert \frac{\Psi^\prime_\gamma \big( \Psi^{-1}_\eta(y) \big) }{\Psi^\prime_\eta \big( \Psi^{-1}_\eta(y) \big)} \Big\vert f_{\Pi} \big( g^{-1}(y) \big)
\end{equation}
Once $f_Y(y)$ has been calculated we use numerical integration to find the expectation.
We find that Gauss-Hermite quadrature \cite{steen1969gaussian} is a sufficiently efficient method for the computation.

\section{Conditions for Decision Convergence} \label{appendixB}
In this section show the existence of decision convergence for experimental setup where the risk preferences of our in-silico agents take values $\eta \in [0,1]$ and where the dynamics of the experiment take values in $\gamma \in [0,1)$.
We will also show the non-existence of decision convergence when $\gamma=1$ as $X_{i,t} \to \infty$.

We first consider the general setup where $\gamma \in [0,1)$.
Let $X_t$ be a discrete-time process with $\Psi_\gamma(x)$ as it's ergodicity transform.
We can then write the evolution of $X_t$ as,
\begin{equation}
    X_{t+1}=\Psi^{-1}_\gamma \big( \Psi_\gamma(X_t)+ Z \big)
\end{equation}
Where $Z$ represents the distribution of the incremental change of the random process $\Psi_\gamma(X_t)$.
The utility function of the growth-rate maximizing agent is $\Psi_\gamma(x)$, therefore the expected change in utility is given by,
\begin{align}
    &\mathbb{E} \bigg( \Psi_\gamma \Big( \Psi^{-1}_\gamma \big( \Psi_\gamma(X_t)+ Z \big) \Big) - \Psi_\gamma(X_t) \Big\vert X_t \bigg) \\
    = &\mathbb{E} \Big( \big( \Psi_\gamma(X_t)+ Z \big) - \Psi_\gamma(X_t) \big\vert X_t \Big) \\
    = & \mathbb{E} (Z)
\end{align}
This shows that the growth-rate maximizing agent will prefer the gamble with the largest $\mathbb{E}(Z)$, that is, the largest expected incremental change of the random process $\Psi_\gamma(X_t)$.

To show decision convergence we wish to show the same preference condition for non growth-rate maximizing agents as $X_t$ diverges from zero.

For non growth-rate maximizing agents we have $U(x)=\Psi_\eta(x)$ with $\eta \neq \gamma$. 
Our approach will be to write the difference in utility as a Taylor series expansion of $Z$, take expectations and show that the coefficient corresponding to the $\mathbb{E}(Z)$ term dominates the higher order terms as $X_t$ diverges from zero.
The Yeo-Johnson transform has the property that $\Psi_\eta(x)=\Psi_{-\eta}(-x)$ hence we only need to consider the case $X_t>0$.

We first define the function $W(z)$ as,
\begin{equation}
    W(z)= \Psi_\eta \Big( \Psi_\gamma^{-1} \big( \Psi_\gamma(X_t)+ z \big) \Big)
\end{equation}
The Taylor series expansion is given by,
\begin{equation}
    W(z)=W(0) + z W^\prime (0) + \frac{z^2}{2} W^{\prime \prime}(0) + \mathcal{O}(z^3)
\end{equation}
As $W(0)$ represents the utility of the agent before the decision is made we have that the expected change in utility is,
\begin{equation} \label{eq16}
    \mathbb{E} \big( W(Z)-W(0) \big\vert X_t \big) = \mathbb{E}(Z) W^\prime (0) + \frac{1}{2} \mathbb{E}(Z^2) W^{\prime \prime}(0) + \mathcal{O}\big( \mathbb{E}(Z^3) \big)
\end{equation}
Using the chain rule,
\begin{align}
    W^\prime(z) &= \Big( (1-\gamma) \big( \Psi_\gamma(X_t) +z\big) \Big)^\frac{1-\eta}{1-\gamma} \\
    \implies W^\prime(0) &\propto \Psi_\gamma(X_t)^\frac{1-\eta}{1-\gamma} 
\end{align}
and therefore the general pattern of higher derivatives is given by,
\begin{equation}
    W^{(n)}(0) \propto \Psi_\gamma(X_t)^{\frac{1-\eta}{1-\gamma}-(n-1)}
\end{equation}
Therefore in \eqref{eq16} as $X_t \to \infty$ the coefficient of $\mathbb{E}(Z)$, that is $W^\prime(0)$, will dominate the coefficients of all higher moment terms.
Therefore the preferences of the agent will converge to preferring the gamble with the largest $\mathbb{E}(Z)$, which is exactly the preferences of the growth-rate maximizing agent.

We now consider the case that $\gamma=1$.
Our approach is the same as before the difference being that now the derivative of $W(z)$ is given by,
\begin{align}
    W^\prime(z) &= \big( (X_t+1) \exp(z) \big)^{1-\eta} \\
    \implies W^\prime(0) &\propto (X_t+1)^{1-\eta} 
\end{align}
and therefore the general pattern of higher derivatives is given by,
\begin{equation}
    W^{(n)}(0) \propto (X_t+1)^{1-\eta}
\end{equation}
The difference in this case is that as $X_t \to \infty$ the coefficient of $\mathbb{E}(Z)$ in \eqref{eq16} no longer dominates the coefficients of higher moments and therefore we do not observe decision convergence when $\gamma=1$.

\bibliography{refs}

\end{document}